# Sparse Additive Functional and Kernel CCA

**Sivaraman Balakrishnan**  SBALAKRI@CS.CMU.EDU
School of Computer Science, Carnegie Mellon University, 5000 Forbes Avenue, Pittsburgh, PA 15213 USA

**Kriti Puniyani**  KPUNIYAN@CS.CMU.EDU
School of Computer Science, Carnegie Mellon University, 5000 Forbes Avenue, Pittsburgh, PA 15213 USA

**John Lafferty**  LAFFERTY@GALTON.UCHICAGO.EDU
Department of Statistics and Department of Computer Science, University of Chicago, Chicago, IL 60637 USA

## Abstract

Canonical Correlation Analysis (CCA) is a classical tool for finding correlations among the components of two random vectors. In recent years, CCA has been widely applied to the analysis of genomic data, where it is common for researchers to perform multiple assays on a single set of patient samples. Recent work has proposed sparse variants of CCA to address the high dimensionality of such data. However, classical and sparse CCA are based on linear models, and are thus limited in their ability to find general correlations. In this paper, we present two approaches to high-dimensional nonparametric CCA, building on recent developments in high-dimensional nonparametric regression. We present estimation procedures for both approaches, and analyze their theoretical properties in the high-dimensional setting. We demonstrate the effectiveness of these procedures in discovering nonlinear correlations via extensive simulations, as well as through experiments with genomic data.

## 1. Introduction

Canonical correlation analysis (Hotelling, 1936), is a classical method for finding correlations between the components of two random vectors $X \in \mathbb{R}^{p_1}$ and $Y \in \mathbb{R}^{p_2}$. Given a set of $n$ paired observations $(X_1, Y_1), \ldots, (X_n, Y_n)$, we form the design matrices $\mathbb{X} \in \mathbb{R}^{n \times p_1}$ and $\mathbb{Y} \in \mathbb{R}^{n \times p_2}$ and find vectors $u \in \mathbb{R}^{p_1}$ and $v \in \mathbb{R}^{p_2}$ that are solutions to the optimization

$$\arg\max_{u,v} \quad \frac{1}{n} u^T \mathbb{X}^T \mathbb{Y} v \qquad (1)$$
$$\text{s.t.} \quad \frac{1}{n} u^T \mathbb{X}^T \mathbb{X} u \leq 1 \qquad \frac{1}{n} v^T \mathbb{Y}^T \mathbb{Y} v \leq 1,$$

where the columns of $\mathbb{X}$ and $\mathbb{Y}$ have been standardized to have mean zero and standard deviation one. This is the sample version of the problem of maximizing the correlation between the linear combinations $u^T X$ and $v^T Y$, assuming the random variables have mean zero.

CCA can serve as a valuable dimension reduction tool, allowing one to quickly zoom in on interesting phenomena shared by multiple data sets. This tool is increasingly attractive in genomic data analysis, where researchers perform multiple assays per item. For instance, data including DNA copy number (or comparative genomic hybridization, CGH), gene expression, and single nucleotide polymorphism (SNP) information can be collected on a common set of patients. Witten et al. (2009) present examples of recent studies involving such data.

When the data are high dimensional, as is often the case for genomic data, the classical formulation of CCA is not meaningful, since the sample covariance matrices $\mathbb{X}^T \mathbb{X}$ and $\mathbb{Y}^T \mathbb{Y}$ are singular. This has motivated different approaches to *sparse* CCA, which regularizes (1) by suitable sparsity-inducing $\ell_1$ penalties (Witten et al., 2009; Witten & Tibshirani, 2009; Parkhomenko et al., 2007; Chen & Liu, 2012). Sparsity can lead to more interpretable models, reduced computational cost, and favorable statistical properties for high dimensional data. Existing methods for CCA are, however, restricted in that they attempt to find linear combinations of the variables—interesting correlations need not be linear. The need for this flexibility motivates the nonparametric approaches we consider in this paper.

The general nonparametric analogue of (1) is

$$\arg\max_{f,g} \frac{1}{n}\sum_{i=1}^{n} f(X_i)g(Y_i) \quad (2)$$
$$\text{s.t.} \quad \frac{1}{n}\sum_{i=1}^{n} f^2(X_i) \leq 1 \quad \frac{1}{n}\sum_{i=1}^{n} g^2(Y_i) \leq 1$$

where $f$ and $g$ are restricted to belong to an appropriate class of smooth functions. Bach & Jordan (2003) introduce a version of this called kernel CCA by applying the "kernel trick" to the CCA problem. Kernel CCA allows flexible nonparametric modeling of correlations, solving (2) with additional regularization to enforce smoothness of the functions $f$ and $g$ in appropriate reproducing kernel Hilbert spaces. However, this general nonparametric model suffers from the curse of dimensionality, as the number of samples required for consistency grows exponentially with the dimension. It is thus necessary to further restrict the complexity of possible functions. We consider the class of additive models which can be written as

$$f(x_1, x_2, \ldots, x_{p_1}) = \sum_{j=1}^{p_1} f_j(x_j) \quad (3)$$

$$g(y_1, y_2, \ldots, y_{p_2}) = \sum_{k=1}^{p_2} g_k(y_k) \quad (4)$$

in terms of univariate component functions (Hastie & Tibshirani, 1986). In the regression setting, such models no longer require the sample size to be exponential in the dimension; however, they only have strong statistical properties in low dimensions. Recently, several authors have shown how sparse additive models for regression can be efficiently estimated even when $p > n$ (Ravikumar et al., 2009; Koltchinskii & Yuan, 2010; Meier et al., 2009; Raskutti et al., 2010).

In this paper we propose two additive nonparametric formulations of CCA, one over a family of RKHSs and another over Sobolev spaces without a reproducing kernel. In the low-dimensional setting where we do not enforce sparsity, the formulation over Sobolev spaces is closely related to the Alternating Conditional Expectations (ACE) formulation of nonparametric regression due to Breiman & Friedman (1985). In addition to formulating algorithms for the optimizations, we provide risk consistency guarantees for the global risk minimizer in the high dimensional regime where $\min(p_1, p_2) > n$.

An important consideration is that sparse nonparametric CCA is biconvex, but not jointly convex in $f$ and $g$. This is true even for the linear CCA model, which is a special case of the model we propose. In the absence of the sparsity constraints the linear problem reduces to a generalized eigenvalue problem which can be efficiently solved. This remains true in the nonparametric case as well. Over an RKHS, the problem without sparsity is a generalized eigenvalue problem where Gram matrices replace the data covariance matrices. In the population setting over the Sobolev spaces we consider, Breiman & Friedman (1985) show that the problem reduces to an eigenvalue problem with respect to conditional expectation operators.

Returning to the nonconvex sparse CCA problem, Witten et al. (2009) and Parkhomenko et al. (2007) suggest using the solution to the nonsparse version of the problem to initialize sparse CCA; Chen & Liu (2012) use several random initializations. As we show in simulations, both approaches can lead to poor results, even in the linear case. To address this issue, we propose and study a simple marginal thresholding step to reduce the dimensionality, in the spirit of the diagonal thresholding of Johnstone & Lu (2009) and the SURE screening of Fan & Song (2010). This results in a three step procedure where after preprocessing we use the nonsparse version of our problem to determine a good initialization for the sparse formulation.

In Sections 2 and 3 we briefly describe the additive Sobolev and RKHS function spaces over which we work, introduce our two nonparametric CCA formulations, and discuss their optimization. In Section 4 we address the non-convexity of the formulations and initialization strategies. In Section 5 we summarize the theoretical guarantees of these procedures when $p_1, p_2 > n$ and in Section 6 we describe some simulations and real data experiments.

## 2. Sparse additive kernel CCA

Recall the linear CCA problem (1). We will now derive its additive generalization over RKHSs. Let $\mathcal{F}_j \subset L_2(\mu(x_j))$ be a reproducing kernel Hilbert space of univariate functions on the domain of $X_j$, and let $\mathcal{G}_k \subset L_2(\mu(y_k))$ be a reproducing kernel Hilbert space of univariate functions on the domain $Y_k$, for each $j = 1, \ldots, p_1$ and $k = 1, \ldots, p_2$. We assume that $\mathbb{E}[f_j(X_j)] = 0$ and $\mathbb{E}[g_k(Y_k)] = 0$ for all $f_j \in \mathcal{F}_j$, and $g_k \in \mathcal{G}_k$ for each $j$ and $k$. This is necessary to enforce model identifiability. In practice, we will always work with centered Gram matrices to enforce this (see Bach & Jordan (2003)).

Denote by $\mathcal{F} = \{f = \sum_{j=1}^{p_1} f_j(x_j) | f_j \in \mathcal{F}_j\}$ and $\mathcal{G} = \{g = \sum_{k=1}^{p_2} g_k(y_k) | g_k \in \mathcal{G}_k\}$ the sets of additive functions of $x$ and $y$, respectively.

We are given $n$ independent tuples of the form $(X_i, Y_i)_{i=1}^n$ where $X_i = \{X_{i1}, \ldots, X_{ip_1}\}$ and $Y_i = \{Y_{i1}, \ldots, Y_{ip_2}\}$, and positive definite kernel functions on each covariate of $X$ and $Y$. We denote the Gram matrix for the $j^{\text{th}}$ X covariate by $K_{xj}$ and for the $k^{\text{th}}$ Y covariate by $K_{yk}$.

We will need to regularize the CCA problem to enforce smoothness and sparsity of the functions. The two norms

$$\|f_j\|_{\mathcal{F}_j} = \sqrt{\langle f_j, f_j\rangle_{\mathcal{F}_j}} \qquad \|f_j\|_2 = \sqrt{\tfrac{1}{n}\sum_{i=1}^n f_j^2(X_{ij})}$$

play an important role in our approach. We can now formulate the *sparse additive kernel CCA* (SA-KCCA) problem as

$$\max_{f \in \mathcal{F}, g \in \mathcal{G}} \frac{1}{n}\sum_{i=1}^n f(X_i)g(Y_i) \quad \text{subject to} \tag{5}$$

$$\frac{1}{n}\sum_{i=1}^n f^2(X_i) + \gamma_f \sum_{j=1}^{p_1} \|f_j\|_{\mathcal{F}_j}^2 \le 1 \quad \sum_{j=1}^{p_1} \|f_j\|_2 \le C_f$$

$$\frac{1}{n}\sum_{i=1}^n g^2(Y_i) + \gamma_g \sum_{k=1}^{p_2} \|g_k\|_{\mathcal{G}_k}^2 \le 1 \quad \sum_{k=1}^{p_2} \|g_k\|_2 \le C_g.$$

for given regularization parameters $\gamma_f, \gamma_g, C_f$ and $C_g$. As with the group LASSO, constraining $\sum_j \|f_j\|_2$ encourages sparsity amongst the functions $f_j$ Ravikumar et al. (2009). As stated, this is an infinite dimensional optimization problem over Hilbert spaces. However, a straightforward application of the representer theorem shows that it is equivalent to the following finite dimensional optimization problem:

$$\max_{\alpha, \beta} \frac{1}{n}\left(\sum_{j=1}^{p_1} K_{xj}\alpha_j\right)\left(\sum_{k=1}^{p_2} K_{yk}\beta_k\right) \quad \text{subject to} \tag{6}$$

$$\frac{1}{n}\left(\sum_{j=1}^{p_1} K_{xj}\alpha_j\right)^T\left(\sum_{j=1}^{p_1} K_{xj}\alpha_j\right) + \gamma_f \sum_{j=1}^{p_1} \alpha_j^T K_{xj}\alpha_j \le 1$$

$$\frac{1}{n}\left(\sum_{k=1}^{p_2} K_{yk}\beta_k\right)^T\left(\sum_{k=1}^{p_2} K_{yk}\beta_k\right) + \gamma_g \sum_{k=1}^{p_2} \beta_k^T K_{yk}\beta_k \le 1$$

$$\sum_{j=1}^{p_1}\sqrt{\tfrac{1}{n}\alpha_j^T K_{xj}^T K_{xj}\alpha_j} \le C_f, \quad \sum_{k=1}^{p_2}\sqrt{\tfrac{1}{n}\beta_k^T K_{yk}^T K_{yk}\beta_k} \le C_g.$$

Here $\alpha$ is an $(n \times p_1)$ matrix, $\alpha_j$ is its $j^{\text{th}}$ column, $\beta$ is an $(n \times p_2)$ matrix and $\beta_k$ is its $k^{\text{th}}$ column.

The problem (6) is not convex. However, if we fix the function $g$ (or equivalently the coefficients $\beta$) the problem is convex in $f$ (equivalently $\alpha$), and vice-versa.

This *biconvexity* leads to a natural optimization strategy for (6) which we describe below. However, this procedure only guarantees convergence to a local optimum and in practice we still need to be able to find a good initialization.

In the absence of the sparsity penalty the problem becomes an additive form of kernel CCA (Bach & Jordan, 2003). One could also consider alternative formulations that, for instance, separate the smoothness and variance constraints. One attractive feature of our formulation is that without the sparsity constraint the problem can be reduced to a generalized eigenvalue computation which can be solved optimally. This leads us to a strategy of biconvex optimization that mirrors the linear algorithm of Witten et al. (2009); specifically, initialize by solving the problem without the sparsity constraints, fix $\alpha$ and optimize for $\beta$ and vice-versa until convergence. As our experiments will show this is indeed a good strategy when $p_1, p_2 < n$. However, new ideas, to be described in Section 4, are necessary to scale this to the high dimensional setting where $p_1, p_2 > n$.

## 3. Sparse additive functional CCA

We now formulate an optimization problem for sparse additive functional CCA (SA-FCCA), and derive a scalable backfitting procedure for this problem. Here we work directly over the Hilbert spaces $L_2(\mu(x))$ and $L_2(\mu(y))$. We will denote by $\mathcal{S}_j$ the subspace of $\mu(x_j)$ measurable functions with mean 0, with the usual inner product $\langle f_j, f_j'\rangle = \mathbb{E}\left(f_j(X_j)f_j'(X_j)\right)$, and similarly $\mathcal{T}_k$ for the functions of $y$.

To enforce smoothness we consider functions lying in a ball in a second order Sobolev space. We further assume the functions are uniformly bounded, and the measures $\mu$ are supported on a compact subset of a Euclidean space with Lebesgue measure $\lambda$. For a fixed uniformly bounded, orthonormal basis $\psi_{jk}$ with respect to $\lambda$ we have

$$\mathcal{F}_j = \Big\{f_j \in \mathcal{S}_j : f_j = \sum_{k=0}^\infty \beta_{jk}\psi_{jk}, \sum_{k=0}^\infty \beta_{jk}^2 k^4 \le C^2\Big\}$$

and similarly for $\mathcal{G}_k$. We will call these the *smooth* functions, and denote by $\mathcal{F}$ and $\mathcal{G}$ the set of smooth additive functions over the respective Hilbert spaces.

Our formulation of *sparse additive functional CCA* is the optimization

$$\max_{f \in \mathcal{F}, g \in \mathcal{G}} \frac{1}{n}\sum_{i=1}^n f(X_i)g(Y_i) \tag{7}$$

s.t. $\frac{1}{n}\sum_{j=1}^{p_1}\sum_{i=1}^{n} f_j^2(X_{ij}) \leq 1, \quad \sum_{j=1}^{p_1} \|f_j\|_2 \leq C_f$

$\frac{1}{n}\sum_{k=1}^{p_2}\sum_{i=1}^{n} g_k^2(Y_{ik}) \leq 1, \quad \sum_{k=1}^{p_2} \|g_k\|_2 \leq C_g$

where the $\|.\|_2$ norm is defined as in additive kernel CCA. This problem is superficially similar to (2); however, there are three important differences. First, we don't regularize for smoothness but instead work directly over a Sobolev space of smooth functions. Secondly, we do not constrain the variance of the function $f$. Instead, in the spirit of "diagonal penalized CCA" of Witten et al. (2009) we constrain the sum of the variances of the individual $f_j$s. This choice is made primarily because it leads to backfitting updates that have a particularly simple and intuitive form. Perhaps most importantly, we can no longer appeal to the representer theorem since we are not working over RKHSs.

We study the population version of this problem to derive a biconvex backfitting procedure to directly optimize this criterion. The sample version of the algorithm is described in Algorithm 1, and a complete derivation is part of the supplementary material. To gain some intuition for this procedure we describe one special case of the population algorithm, where $g$ is fixed and both constraints on $f$ are tight. Consider the Lagrangian problem

$$\max_{f} \min_{\lambda \geq 0, \gamma \geq 0} \mathbb{E}[f(X)g(Y)] - \lambda(\|f\|_2^2 - 1) - \gamma(\|f\|_1 - C_f).$$

The norms are defined as $\|f\|_1 = \sum_{j=1}^{p_1} \sqrt{\mathbb{E}(f_j^2(x_j))}$ and $\|f\|_2^2 = \sum_{j=1}^{p_1} \mathbb{E}(f_j^2(x_j))$. For simplicity, consider the case when $\lambda, \gamma > 0$, and denote $a \equiv g(Y)$.

We now can derive a coordinate ascent style procedure where we optimize over $f_j$ holding the other functions fixed. The Fréchet derivative w.r.t. $f_j$ in the direction $\eta$ gives one of the KKT conditions $\mathbb{E}[(a - 2\lambda f_j - \gamma \nu_j)\eta] = 0$ for all $\eta$ in the Hilbert space $\mathcal{H}_j$, where the subdifferential is $\nu_j = \frac{f_j}{\sqrt{\mathbb{E}(f_j^2)}}$ if $\sqrt{\mathbb{E}(f_j^2)}$ is not 0, and is the set $\{u_j \in \mathcal{H}_j \mid \mathbb{E}(u_j^2) \leq 1\}$ if $\sqrt{\mathbb{E}(f_j^2)} = 0$.

Using iterated expectations the KKT condition can be written as $\mathbb{E}[(\mathbb{E}(a \mid X_j) - 2\lambda f_j - \gamma \nu_j)\eta] = 0$. Denote $E(a \mid X_j) \equiv P_j$. In particular, if we consider $\eta = \mathbb{E}[(\mathbb{E}(a \mid X_j) - 2\lambda f_j - \gamma \nu_j]$, we can see that $\mathbb{E}[(\mathbb{E}(a \mid X_j) - 2\lambda f_j - \gamma \nu_j)] = 0$, i.e., $\mathbb{E}(a \mid X_j) - 2\lambda f_j - \gamma \nu_j = 0$ almost everywhere.

Then if $\sqrt{\mathbb{E}(P_j^2)} \leq \gamma$, we have $f_j = 0$, and we arrive at the following soft thresholding update:

$$f_j = \frac{1}{2\lambda}\left[1 - \frac{\gamma}{\sqrt{\mathbb{E}(P_j^2)}}\right]_+ P_j.$$

Now, going back to the constrained version, we need to select $\gamma$ and $\lambda$ so that the two constraints are tight. To get the sample version of this update we replace the conditional expectation $P_j$ by an estimate $S_j a$, where $S_j$ is a locally linear smoother.

---

**Algorithm 1** Biconvex backfitting for SA-FCCA

**input** $\{(X_i, Y_i)\}$, parameters $C_f, C_g$, initial $g(Y_i)$

1. Compute smoothing matrices $S_j$ and $T_k$.
2. Fix $g$. For each $j$, set $f_j \leftarrow \frac{S_j g}{\lambda}$ where $\lambda = \sqrt{\sum_{j=1}^{p_1}(g^T S_j^T S_j g)}$
3. **if** $\sum_{j=1}^{p_1} \|f_j\|_2 \leq C_f$ , break
   **else** let $\mathcal{F}_m$ denote the functions with maximum $\|.\|_2$ norm. Set all other functions to 0. For each $f \in \mathcal{F}_m$, set $f \leftarrow \frac{C_f f}{|\mathcal{F}_m| \|f\|_2}$. If $\sum_{j=1}^{p_1} \|f_j\|_2^2 \leq 1$, break
   **else** set $f_j \leftarrow \left(1 - \frac{\gamma}{\sqrt{\|S_j g\|_2}}\right)_+ \frac{S_j g}{\lambda}$ where $\lambda = \sqrt{\sum_{j=1}^{p_1}\left\|\left(1 - \frac{\gamma}{\sqrt{\|S_j g\|_2}}\right)_+ S_j g\right\|_2^2}$ and $\gamma$ is chosen so that $\sum_{j=1}^{p_1} \sqrt{g^T S_j^T S_j g} = C_f$
4. Center by setting each $f_j \leftarrow f_j - \text{mean}(f_j)$.
5. Fix $f$ and repeat above to update $g$. Iterate both updates till convergence.

**output** Final functions $f$, $g$

---

## 4. Marginal Thresholding

The formulations of SA-KCCA and SA-FCCA above are not jointly convex, but are biconvex. Hence, iterative optimization algorithms may not be guaranteed to reach the globally optimal solution. To address this issue, we first run the algorithms without any sparsity constraint. The resulting nonsparse collections of functions are then used as initializations for the algorithm that incorporates the sparsity penalties. While such initialization works well for low dimensional problems, as $p$ increases, the performance of the estimator goes down (Figure 1). To extend the algorithms to the high dimensional scenario, we propose marginal thresholding as a screening method to reject irrelevant variables and run the SA-FCCA and SA-KCCA models on the reduced dimensionality problem. For each pair of variables $X_i$ and $Y_j$, we fit marginal functions to that pair by optimizing the criteria in either Equation (6) or Equation (7) *without* the sparsity constraints since we only consider one $X$ and one $Y$ covariate at a time. We then compute the correlation on held out data. This constructs a matrix $M$ of size $p_1 \times p_2$ with $(i, j)$ entry of the matrix representing an estimate of the marginal correlation between $f_i(X_i)$ and $g_j(Y_j)$. We then threshold the entries of $M$ to obtain a subset of variables on which to run SA-FCCA and SA-KCCA. Theorem 5.3 discusses the theoretical properties of marginal thresholding as a screening procedure, and Section 6.2 presents results on marginal thresholding for high dimensional problems.

| Init | p=10 | p=25 | p=50 |
|---|---|---|---|
| Random | 0.05 | 0.009 | -0.02 |
| Non-sparse | **0.97** | 0.62 | 0.26 |

Table 1. Test correlation from functions estimated by SA-FCCA for $n = 75$ samples, where $Y_1 = X_1^2$, all other dimensions are Gaussian noise. Random initializations don't work well for all data sizes. Initializing with the non-sparse formulation works well when $n > p$, but fails as $p \geq n$.

## 5. Main theoretical results

In this section we will characterize both the functional and kernel marginal thresholding procedures and study the theoretical properties of the estimators (6) and (7). We will state the main theorems and defer all proofs to the supplementary material.

The theoretical characterization of these procedures relies on *uniform* large deviation inequalities for the covariance between functions. For simplicity in this section we will assume all the univariate spaces are identical. In the RKHS case we restrict our attention to functions in a ball of a *constant* radius in the Hilbert space associated with a reproducing kernel $K$. In the functional case the univariate space is a second order Sobolev space where the integral of the square of the second derivative is bounded by a *constant*. With some abuse of notation we will denote these spaces $\mathcal{C}$. We are interested in controlling the quantity

$$\Theta_n = \sup_{f_j, g_k} \left| \frac{1}{n} \sum_{i=1}^n f_j(X_{ij}) g_k(Y_{ik}) - \mathbb{E}(f_j(X_j) g_k(Y_k)) \right|$$

where $f_j, g_k \in \mathcal{C}, j \in \{1, \ldots, p_1\}, k \in \{1, \ldots, p_2\}$.

All results extend to the case when each covariate is endowed with a possibly distinct function space.

**Lemma 5.1 (Uniform bound over RKHS)**
*Assume* $\sup_x |K(x,x)| \leq M < \infty$, *for functions* $f_j(x) = \sum_{i=1}^n \alpha_{ij} K_x(x, X_{ij})$, $g_k(y) = \sum_{i=1}^n \beta_{ik} K_y(y, Y_{ik})$

$$\mathbb{P}\left( \Theta_n \geq \underbrace{\zeta + C\sqrt{\frac{\log((p_1 p_2)/\delta)}{n}}}_{\epsilon} \right) \leq \delta$$

*where $C$ is a constant depending only on $M$, and $\zeta = \max_{j,k} \frac{2}{n} \mathbb{E}_{X \sim x_j, Y \sim y_k} \sqrt{\sum_{i=1}^n K(X_{ij}, X_{ij}) K(Y_{ik}, Y_{ik})}$*

Note that $\zeta$ is independent of the dimensions $p_1$ and $p_2$ and that under the assumption that $K$ is bounded, $\zeta = O(1/\sqrt{n})$. In some cases however this term can be much smaller. The second term depends only logarithmically on $p_1$ and $p_2$ and this *weak* dependence is the main reason our proposed procedures are consistent even when $p_1, p_2 > n$.

**Lemma 5.2 (Uniform bound for Sobolev spaces)**
*Assume* $\|f\|_\infty \leq M \leq \infty$, *then*

$$\mathbb{P}\left( \Theta_n \geq \underbrace{\frac{C_1}{\sqrt{n}} + C_2 \sqrt{\frac{\log((p_1 p_2)/\delta)}{n}}}_{\epsilon} \right) \leq \delta$$

*where $C_1$ and $C_2$ depend only on $M$.*

Lemma 5.1 is proved via a Rademacher symmetrization argument of Bartlett & Mendelson (2002) (see also Gretton et al. (2004)) while Lemma 5.2 is based on a bound on the bracketing integral of the Sobolev space (see Ravikumar et al. (2009)). The Rademacher bound gives a distribution dependent bound which can in some cases lead to faster rates.

We are now ready to characterize the marginal thresholding procedure described in Section 4. To study marginal thresholding we need to define *relevant* and *irrelevant* covariates. For each covariate $X_j$, denote

$$\alpha_j = \sup_{f_j, g_k \in \mathcal{C}, k \in \{1, \ldots, p_2\}} \mathbb{E}(f_j(X_j) g_k(Y_k))$$

with $\mathbb{E}(f_j^2) \leq 1, \mathbb{E}(g_k^2) \leq 1$. A covariate $X_j$ is considered irrelevant if $\alpha_j = 0$ and relevant if $\alpha_j > 0$. Similarly, for each $Y_k$ we associate $\beta_k$ defined analogously.

Now, assume that for every pair of covariates, we find the maximizer of the SA-FCCA or SA-KCCA objective over the given sample, over the appropriate class $\mathcal{C}$ and with $\mathbb{E}(f_j^2) \leq 1, \mathbb{E}(g_k^2) \leq 1$. Recall that for marginal thresholding we do not enforce sparsity. The global maximization of the SA-KCCA objective can be efficiently carried out since it is equivalent to a generalized eigenvalue problem. For SA-FCCA however, the backfitting procedure is only guaranteed to find the global maximizer in the population setting.

**Theorem 5.3** *Given* $\mathbb{P}(\Theta_n \geq \epsilon) \leq \delta$.

1. *With probability at least* $1-\delta$, *marginal thresholding at* $\epsilon$ *has no false inclusions.*
2. *Further, if we have that* $\alpha_j$ *or* $\beta_k \geq 2\epsilon$ *then under the same* $1-\delta$ *probability event marginal thresholding at* $\epsilon$ *correctly includes the relevant covariate* $X_j$ *or* $Y_k$.

The importance of Lemmas 5.1 and 5.2 is that they provide values at which to threshold the marginal covariances. In particular, notice that the minimum sample covariance that can be reliably detected, with no false inclusions, falls rapidly with $n$ and approaches zero even when $p_1, p_2 > n$.

In the spirit of early results on the LASSO of Juditsky & Nemirovski (2000); Greenshtein & Ritov (2004) we will establish the risk consistency or *persistence* of the empirical maximizers of the two objectives. Although we cannot guarantee that we find these empirical maximizers due to the non-convexity this result shows that with good initialization the formulations (6) and (7) can lead to solutions which have good statistical properties in high dimensions.

For SA-KCCA we will assume that our algorithm maximizes

$$\frac{1}{n}\sum_{i=1}^{n}\left[\sum_{j=1}^{p_1}\mu_j f_j(X_{ij})\right]\left[\sum_{k=1}^{p_2}\gamma_k g_k(Y_{ik})\right]$$

over the classes

$$\mathcal{F} = \left\{f : f(x) = \sum_{j=1}^{p_1}\mu_j f_j(x_j), \mathbb{E}f_j = 0, \mathbb{E}f_j^2 = 1,\right.$$

$$\left.\|\mu\|_1 \leq C_f, \|\mu\|_2^2 + \gamma_f \sum_{j=1}^{p_1}\|f_j\|_{\mathcal{H}}^2 \leq 1\right\}$$

$$\mathcal{G} = \left\{g : g(x) = \sum_{k=1}^{p_2}\gamma_k g_k(y_k), \mathbb{E}g_k = 0, \mathbb{E}g_k^2 = 1,\right.$$

$$\left.\|\gamma\|_1 \leq C_g, \|\gamma\|_2^2 + \gamma_g \sum_{k=1}^{p_2}\|f_k\|_{\mathcal{H}}^2 \leq 1\right\}$$

and for SA-FCCA we will assume that our algorithm maximizes the same objective over the same class without the RKHS constraint but which are instead in a Sobolev ball of constant radius. Denote these solutions $(\hat{f}, \hat{g})$.

We will compare to an *oracle* which maximizes the population covariance

$$\text{cov}(f, g) \equiv \mathbb{E}\left[\sum_{j=1}^{p_1}\mu_j f_j(x_j)\right]\left[\sum_{k=1}^{p_2}\gamma_k g_k(y_k)\right]$$

Denote this maximizer by $(f^*, g^*)$. Our main result will show that these procedures are *persistent*, i.e., $\text{cov}(f^*, g^*) - \text{cov}(\hat{f}, \hat{g}) \to 0$ even if $p_1, p_2 > n$.

**Theorem 5.4 (Persistence)** *If* $p_1 p_2 \leq e^{n^{\xi}}$ *for some* $\xi < 1$ *and* $C_f C_g = o(n^{(1-\xi)/2})$, *then SA-FCCA and SA-KCCA are persistent over their respective function classes.*

## 6. Experiments

### 6.1. Non-linear correlations

We compare SA-FCCA and SA-KCCA with two models, sparse additive linear CCA (SCCA) (Witten et al., 2009) and kernel CCA (KCCA) (Bach & Jordan, 2003). Figure 1 shows the performance of each model, when run on data with $n = 150$ samples in $p_1 = 15$, $p_2 = 15$ dimensions, where only one relevant variable is present in $X$ and $Y$ (the remaining dimensions are Gaussian random noise). We report two metrics to measure whether the correct correlations are being captured by the different methods - (a) test correlation on 200 samples, using the estimated functions, and (b) precision and recall in identifying the correct variables involved in the correlation estimation. Each result is averaged over 10 repeats of the experiment. Since KCCA uses all data dimensions in finding correlations, its precision and recall are not reported.

When the relationship between the relevant variables is linear, all methods identify the correct variables and have high test correlation. While KCCA should be able to identify non-linear correlations, since it is strongly affected by the curse of dimensionality, it has poor test correlation even in $p = 15$ dimensions.

Both SA-FCCA and SA-KCCA correctly identify the relevant variables in all cases, and have high test correlation.

### 6.2. Marginal thresholding

We now test the efficiency of marginal thresholding by running an experiment for $n = 150$, $p_1 = 150$, $p_2 = 150$. We generate multiple relevant variables as:

$$f_i(X_i) = \cos\left(\frac{\pi}{2}X_i\right), \ i \in \{1, 3\}, \ f_i(X_i) = X_i^2, \ i \in \{2, 4\}$$

$$Y_j = \sum_{i=1; i \neq j}^{4} f_i(X_i) + \mathcal{N}(0, 0.1^2) \quad j \in \{1, 2, 3, 4\}$$

Thus, there are four relevant variables in each data set. $X$ and $Y$ are sampled from a uniform distribution, and standardized before computing $f_i(X_i)$. Each $f_i(X_i)$ is

# Sparse Additive Functional and Kernel CCA

| Model | Test correlation | | | | Precision/Recall | | |
|---|---|---|---|---|---|---|---|
| | SA-FCCA | SA-KCCA | SCCA | KCCA | SA-FCCA | SA-KCCA | SCCA |
| $Y = X^2$ | 0.96 | **0.99** | 0.05 | 0.44 | 1/1 | 1/1 | 0.28/0.14 |
| $Y = \text{abs}(X)$ | **0.98** | 0.99 | 0.06 | 0.35 | 1/1 | 1/1 | 0/0 |
| $Y = \cos(X)$ | 0.94 | **0.99** | 0.071 | 0.04 | 1/1 | 1/1 | 0.1/0.1 |
| $\log(Y) = \sin(X)$ | 0.91 | **0.93** | 0.22 | 0.09 | 1/1 | 1/1 | 0.71/0.66 |
| $Y = X$ | **0.99** | 0.99 | 0.99 | 0.98 | 1/1 | 1/1 | 1/1 |

*Figure 1.* Test correlations, and precision and recall for identifying relevant variables for the four different methods. SA-FCCA and SA-KCCA find strong correlations in the data, in both linear and non-linear settings. In all five data sets, SA-FCCA and SA-KCCA are always able to find the relevant variables.

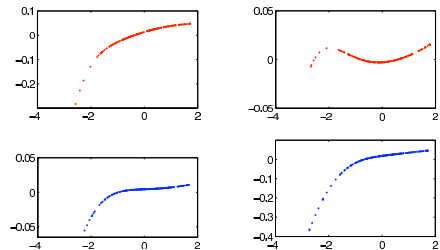

*Figure 2.* DLBCL data : The top row shows two of the functions $f_i(X_i)$ with non-zero norms for $X$ in red, and the bottom row shows two functions $g_j(Y_j)$ with non-zero norms for $Y$ in blue.

also standardized before computing $Y_j$. We repeat the experiment by generating data 10 times, and report results in Table 2. Bandwidth in the different methods was selected using a plug-in estimator of the median distance between points in a single dimension. The sparsity and smoothness parameters for all methods were tuned using permutation tests, as described in Witten et al. (2009), assuming that $C_f = C_g = C$, and $\gamma_f = \gamma_g = \gamma$.

We ran marginal thresholding by splitting the data into equal sized train and held out data, fitting marginal functions on the train data, computing functional correlation on the held out data, and picking a threshold so that $n/5$ elements of the thresholded correlation matrix are non-zero. We found that in all experiments, marginal thresholding always selected the relevant variables for the subsampled data. Table 2 shows the precision, recall and test correlations for the different methods. As can be expected, SA-FCCA and SA-KCCA are able to correctly identify the relevant variables, and the estimated functions have high correlation on test data.

We visualize the effect of the parameter tuning by plotting regularization paths, as the sparsity parameter is varied ($n=100$, $p_1=p_2=12$). For SA-FCCA and SA-KCCA, the norm of each function is plotted, and for sparse linear CCA, the absolute values of the entries of $u$ and $v$ are shown. Figure 3 shows how, unlike SCCA, SA-FCCA and SA-KCCA are able to separate the relevant and non-relevant variables over the entire range of the sparsity parameter.

## 6.3. Application to DLBCL data

We apply our non-linear CCA models to a data set of comparative genomic hybridization (CGH) and gene expression measurements from 203 diffuse large B-cell lymphoma (DLBCL) biopsy samples (Lenz, 2008). We obtained 1500 CGH measurements from chromosome

| Method | Test correlation | Precision | Recall |
|---|---|---|---|
| SA-FCCA | **0.94** | 1 | 0.785 |
| SA-KCCA | **0.98** | 0.95 | 0.8 |
| SCCA | 0.02 | 0.02 | 0.36 |
| KCCA | 0.07 | N/A | N/A |

*Table 2.* Test correlations, precision and recall for identifying the correct relevant variables for the four different methods ($n = 150$, $p_1 = 150$, $p_2 = 150$). Marginal thresholding was used for selecting relevant variables before running SA-FCCA and SA-KCCA

1 of the data, and 1500 gene expression measurements from genes on chromosome 1 and 2 of the data. The data was standardized,and Winsorized so that the data lies within two times the mean absolute deviation.

We used marginal thresholding to reduce the dimensionality of the problem, and then ran SA-FCCA. Permutation tests were used to pick an appropriate bandwidth and sparsity parameter, as described in Witten et al. (2009). We found that the model picked interesting non-linear relationships between CGH and gene expression data. Figure 2 shows the functions extracted by the SA-FCCA model from this data. Even though this data has been previously analyzed using linear models, we do not necessarily expect gene expression measurements from Affymetrix chips to be linearly correlated with array CGH measurements, even if the specific CGH mutation is truly affecting the gene expression. Further, the extracted functions in Figure 2 suggest that the changes in gene expression are dependent on the CGH measurements via a saturation function - as the copy number increases, the gene expression increases, until it saturates to a fixed level, beyond which increasing the copy numbers does not lead to an increase in expression. From a systems biology view point, such a prediction seems reasonable since single CGH mutations will not affect other pathways that are required to be activated for large changes in gene expression.

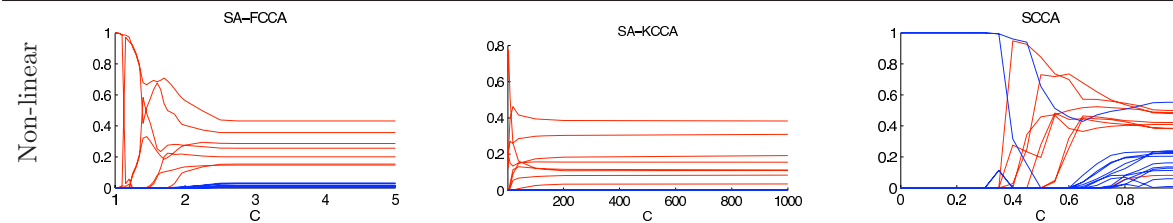

Figure 3. Regularization paths for non-linear correlations in the data, for SA-FCCA, SA-KCCA and SCCA resp. The paths for the relevant variables (in $X$ and $Y$) are shown in red, the irrelevant variables are shown in blue.

## 7. Discussion

In this paper we introduced two proposals for nonparametric CCA and demonstrated their effectiveness both in theory and practice. Several interesting questions and extensions remain. CCA is often run on more than two data sets, and one is often interested in more than just the *principal* canonical direction. Chen & Liu (2012) have proposed group sparse linear CCA for situations when a grouping of the covariates is known. These extensions all have natural nonparametric analogues which would be interesting to explore. As in the case of regression (Koltchinskii & Yuan, 2010), the KCCA formulation considered in this paper can also be generalized to involve multiple kernels and kernels over groups of variables in a straightforward way.

While thresholding marginal correlations one can imagine exploiting the structure in the correlations. In particular, in the $(p_1 \times p_2)$ marginal correlations matrix we are looking for a *bicluster* of high entries in the matrix. Leveraging this structure could potentially allow us to detect weaker marginal correlations. Finally, an important application of kernel CCA is as a contrast function in independence testing. The additive formulations we have proposed allow for independence testing over more restricted alternatives but can be used to construct *interpretable* tests of independence.

## Acknowledgements

Research supported in part by NSF grant IIS-1116730, AFOSR contract FA9550-09-1-0373, and NIH grant R01-GM093156-03.